\pdfoutput=1

\documentclass[11pt]{article}

\usepackage{latex/acl}

\usepackage{times}
\usepackage{multirow}
\usepackage{latexsym}
\usepackage{graphicx}
\usepackage{mathletters}
\usepackage{float}

\usepackage[T1]{fontenc}
\usepackage{enumitem}
\usepackage{amssymb}
\usepackage{mathtools}
\usepackage{amsthm}
\usepackage{amsfonts}
\usepackage{dsfont}
\usepackage{subfig}
\usepackage{caption}
\usepackage{subcaption}
\usepackage{eurosym,rotating,booktabs}
\usepackage{amsmath}
\usepackage{algorithm2e}
\RestyleAlgo{ruled}
\usepackage[capitalize,noabbrev]{cleveref}
\usepackage{devanagari}
\usepackage[english]{babel} 
\babelprovide[import]{hindi}
\babelprovide[import]{sanskrit}
\theoremstyle{plain}

\theoremstyle{definition}

\theoremstyle{remark}

\usepackage[utf8]{inputenc}

\usepackage{microtype}

\usepackage{inconsolata}

\title{Isometric Neural Machine Translation using Phoneme Count Ratio Reward-based Reinforcement Learning}

 \author{Shivam Ratnakant Mhaskar$^1$, Nirmesh J. Shah$^1$, Mohammadi Zaki$^1$, \\ {\bf Ashishkumar P. Gudmalwar$^1$,  
 Pankaj Wasnik$^1$, Rajiv Ratn Shah $^2$} \\ $^1$ Sony Research India, Bangalore \\$^2$Indraprastha Institute of Information Technology (IIIT), Delhi \\   \text{\{nirmesh.shah;mohammadi.zaki;ashish.gudmalwar1;pankaj.wasnik\}@sony.com,}\\ \text{rajivratn@iiitd.ac.in}}

\begin{document}
\maketitle

\begin{abstract}

Traditional Automatic Video Dubbing (AVD) pipeline consists of three key modules, namely, Automatic Speech Recognition (ASR), Neural Machine Translation (NMT), and Text-to-Speech (TTS). Within AVD pipelines, isometric-NMT algorithms are employed to regulate the length of the synthesized output text. This is done to guarantee synchronization with respect to the alignment of video and audio subsequent to the dubbing process.
Previous approaches have focused on aligning the number of characters and words in the source and target language texts of Machine Translation models. However, our approach aims to align the number of phonemes instead, as they are closely associated with speech duration. In this paper, we present the development of an isometric NMT system using Reinforcement Learning (RL), with a focus on optimizing the alignment of phoneme counts in the source and target language sentence pairs. To evaluate our models, we propose the Phoneme Count Compliance (PCC) score,  which is a measure of length compliance.
Our approach demonstrates a substantial improvement of approximately \textbf{36\%} in the PCC score compared to the state-of-the-art models when applied to English-Hindi language pairs. Moreover, we propose a student-teacher architecture within the framework of our RL approach to maintain a trade-off between the phoneme count and translation quality. 

\end{abstract}

\section{Introduction}
Automatic Video Dubbing (AVD) technologies have become popular in recent times with the advent of Generative AI technologies. AVD technology automatically converts a video from one language to another language in three steps, \textit{(i)} Automatic Speech Recognition (ASR) \textit{(ii) }Neural Machine Translation (NMT), and \textit{(iii)} Text-to-Speech (TTS). This task has become crucial especially in content creation as it helps to break down language barriers and reach a wider audience.   
A crucial factor underlying the quality and effectiveness of an AVD system is the synchronization of the audio and video post-dubbing. For seamless and consistent synchronization, the duration of the target language speech generated by TTS in the AVD system must match with the duration of the source language speech. If the duration is not matched, various signal processing techniques can be applied to a certain extent to manipulate the duration of the final audio. However, this process introduces artifacts and degrades the quality of TTS output. Hence, a major focus of the research community has shifted towards controlling the length of the text output after NMT, such that there is much less mismatch in duration after dubbing. In this paper, we strive to enhance the performance of the Isometric NMT model, introduced in \cite{lakew2022isometric}, which is tasked with controlling the length of generated texts. 

Traditionally machine translation for AVD has been done as a two-step process \cite{lakew21verbosity}, where for every input sentence, various output sentences are generated and then re-ranked according to length-matching. \cite{lakew2022isometric} marked the advent of self-learning methods for the NMT task for AVD.
Further works aimed to produce output texts with the duration compliance directly \cite{wu2023videodubber}. 
However, these models rely on training a separate duration generation model for the length compliance, which is computationally too expensive. Furthermore, works like \cite{lakew-etal-2019-controlling} use the matching of the number of characters or words between the source and target language sentences. However, in this work, we model this problem as matching the number of \textit{phonemes} between the source and target language sentences because phonemes have a closer association with the speech duration \cite{quatieri, oppenheim99}. We model this \textit{matching} as a reward indicator which simplifies and speeds up the training process in contrast with some previous works \cite{wu2023videodubber} where the duration of translated texts was controlled using estimates of phoneme lengths, which is time-consuming \cite{wu2023videodubber}. 

In addition, we propose a Reinforcement Learning (RL) based training strategy to achieve the task of isometric NMT in the context of generating translation outputs such that the phoneme counts of the source and target language sentences are as close as possible. We first translate the source language sentences using a pre-trained transformer-based NMT model (generation step), which we treat as an RL agent. Then, we compute the ratio between the phoneme counts of the source and the generated target language sentences. After this, we filter out sentences where the phoneme count ratio (PCR) deviates from a pre-defined threshold determined empirically. We then use the filtered data for finetuning the agent model. We perform multiple iterations of generation using the RL agent and subsequent finetuning using the duration-based positively rewarded dataset. 


With each finetuning step, we make the PCR criteria stricter by increasing the threshold value for reward strategies, which positively reflects in the results we obtain (see Sec.\ref{sec:results}), by achieving higher PCC scores. However, this adversely affects the translation quality. To address this issue, we modify the RL-agent (i.e., a fine-tuned (FT) model) with the help of knowledge distillation step via a student-teacher architecture (see Figure \ref{fig:rl}) \cite{hinton2015distilling}. We use knowledge-distillation step and Student-teacher interchangeably throughout the paper. Here, the \emph{teacher} is the SOTA NMT model (i.e., the model with the best BLEU score, but, possibly a poor PCC score). This further finetuning step helps the \emph{student} (the current FT-model) to learn to produce good-quality as well as phoneme count compliant output. The effectiveness of our proposed model is demonstrated for the English-Hindi language pair (Hindi is spoken by more than 500 million people). For the training and validation, we used the BPCC corpus \cite{gala2023indictrans2}, and for testing, we used i) held-out BPCC Test corpus, ii) Flores, and iii) a movie database (see Section 4.1). We significantly improved the performance of the English-Hindi NMT with respect to various metrics like BLEU, BLEURT, COMET, chrF, and a novel metric, namely, PCC which measures the length compliance between the source and translated sentence. \\ 
We summarize our contributions as:
\begin{enumerate}[wide, labelwidth=!, labelindent=0pt]
    \item  To the best of our knowledge this is the first attempt to apply a RL strategy for achieving Isometric NMT.
    \item We propose a method to match phoneme \textit{counts} in source and target sentences to control duration using a reward strategy in RL, aiming to enhance synchronization in the AVD task.
    \item To address translation quality degradation from constrained duration in source and target language translations, we propose a student-teacher architecture as a post-processing step for the RL-NMT approach. 
    \item The work centers on AVD for English-to-Hindi languages, an area that has been relatively neglected until now. 
    \item We benchmark the performance of our proposed approaches against many state-of-the-art models and Large Language Models (LLMs).
\end{enumerate}
The paper is structured as follows. Section 2 discusses related work. Section 3 discusses in detail the methodology. Section 4 presents details of experiments and results. Section 5 concludes the paper and presents limitations of the work. 
\section{Related Work}
Neural Machine Translation models \cite{bahdanau2014neural,cho-etal-2014-properties, sutskever2014sequence} have majorly improved the performance in the machine translation task. Transformer \cite{vaswani2017attention} architecture is widely used in state-of-the-art NMT models. Automatic Video Dubbing pipeline requires the use of NMT models which produce outputs such that the corresponding speech duration of the target language sentence matches the speech duration of the source language sentence. \citet{lakew-etal-2019-controlling} formulated this problem as matching the number of characters in the source and target language sentences. They injected the information regarding the number of characters in the positional embeddings with the help of tags appended to the source language sentence. 
In the work, \citet{lakew2022isometric} introduced a self-training approach, both offline and online, and implemented the tagging of source sentences with the length ratio between the source and target language sentences, calculated based on the number of characters. \citet{wu2023videodubber} formulated the problem as matching the duration in terms of the number of mel-frames of the source and target language sentences. They incorporated the number of mel-frames in positional embeddings of the transformer architecture.
\section{Methodology}
In this section, we first discuss the problem setup of formulating the  MT task in the RL framework. Next, we propose our RL-based training approach for Isometric NMT for achieving phoneme count compliant translation. Finally, we conclude by proposing the student-teacher architecture by modifying the agent in the RL training to mitigate the problem of quality degradation.
\subsection{Problem Setup}
The machine translation (MT) task can be cast into a Reinforcement Learning (RL) problem \cite{ReSTDeepMind}.\footnote{We push the detailed Markov Decision Process (MDP) formalism to the appendix, due to space constraints.} We consider the problem of translating an input sentence $\mathbf{x}$ from a source language $A$ to sentence $\mathbf{y}$ in some other target language $B$. To integrate the MT task into an automatic dubbing pipeline, we strive towards generating the output sentence $\mathbf{y}$ to have (nearly) the same number of phonemes as the input sentence $\mathbf{x}$, which would imply better duration alignment between the input and the output languages.

Let the input and the output (target) sentences consist of $n$ and $m$ tokens (words/sub-words, etc.), respectively. Then, with some abuse of notation, a machine translation system characterized by a policy $p$, which takes as input a sequence of vectors ${\bf{x}}\equiv(x_1,x_2,\ldots,x_n)$ (where each $x_i$ is an embedding vector according to the input vocabulary) and generates an output sequence of vectors $\mathbf{y}\equiv(y_1,y_2,\ldots, y_m)$ can be expressed as an auto-regressive product of the probability distribution using the Bayes' Theorem as shown in Eq. \ref{eq: language model},
\begin{equation} \label{eq: language model} 
p(\mathbf{y}\given \mathbf{x},w) = \prod_{s=1}^m p(y_s\given y_1,\ldots, y_{s-1},\mathbf{x},w),    
\end{equation}
 where $w$ are the parameters defining the policy. For the automatic dubbing task, to enforce the importance of the equal time duration of the input and output texts, we define a notion of a reward $r(.,.)$ as a function that takes two arguments, namely, $\hat{\mathbf{y}}$ and $\mathbf{x}$. Here $\hat{\mathbf{y}}$ is the translated sentence for the input sentence $\mathbf{x}$ by the system. Then $r(\hat{\mathbf{y}},\mathbf{x})$ is chosen as a function of the Phoneme Count Ratio (PCR) score. In particular, for some (small) $\delta>0$ we set as shown in Eq. \ref{eq: reward formulation},
\begin{equation}\label{eq: reward formulation}
r(\hat{\mathbf{y}},\mathbf{x}):=\ind\left\{PCR(\hat{\mathbf{y}},\mathbf{x})\in [1-\delta,1+\delta]\right\}.    
\end{equation}
We aim to optimize the following blend of the two loss (reward) functions (see Eq. \ref{eq: optimization problem}), which would help achieve good translation quality along with reasonable time-duration compliance between the input and the output texts,  
\begin{equation}\label{eq: optimization problem}
    \max\limits_w-\mathbb{E}_{\mathbf{x}\sim \cD}\left[r(\hat{\mathbf{y}},\mathbf{x})\left(\sum\limits_{s=1}^M \log p(\hat{y}_s| \hat{y}_{<s},\mathbf{x}, w)\right) \right].
\end{equation}





\subsection{Proposed Reinforcement Learning based Training for Isometric NMT (RL-NMT)}
\begin{figure*}[h]
    \centering
    \includegraphics[width=\textwidth]{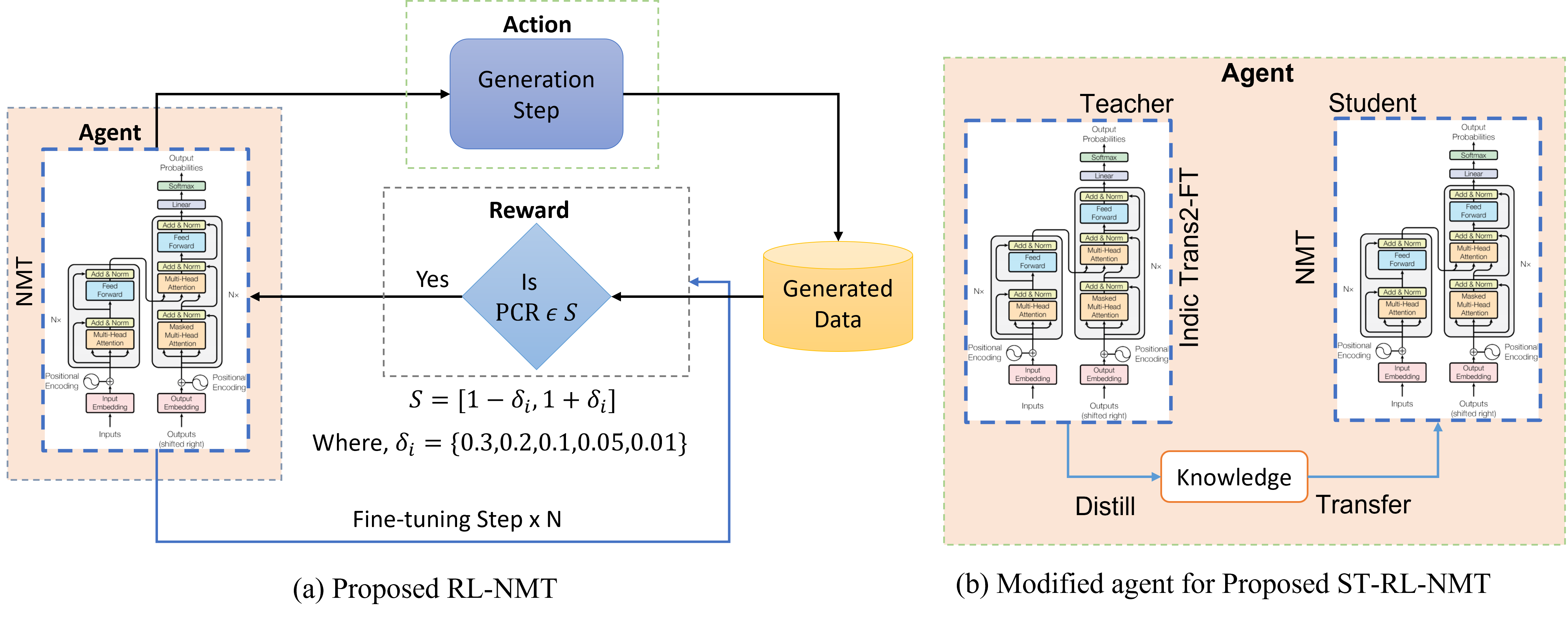}
    \vspace{-0.7cm}
    \caption{Schema showing (a) block diagram of the proposed RL-NMT architecture  (b) modified agent with student-teacher (ST) framework for quality-duration balance.}
    \label{fig:rl}
    \vspace{-0.5cm}
\end{figure*}
\begin{algorithm*}[h]
    \caption{Reinforcement Learning based training algorithm for Isometric NMT}\label{alg:RLNMT}
    \textbf{Terminologies:} $x$: source language sentence, $y$: target language sentence, $\hat{y}$: target language sentence produced by NMT model $\mathcal{M}$, $\mathcal{D}_G$: Generated Dataset, $\mathcal{D}_F$: Filtered Dataset, $N$: Number of parallel sentences in dataset $\mathcal{D}$ \\ 
    \textbf{Input:} $\mathcal{D}$: Dataset,  $\mathcal{M}$: Initial NMT model, $\mathcal{L}_{CE}$: Cross Entropy Loss, $\mathcal{L}_{KL}$: KL Divergence Loss, $\mathcal{L}$: Overall Loss ($\mathcal{L} = \mathcal{L}_{CE} + \alpha \mathcal{L}_{KL}$), $\mathcal{G}$: Number of Generation steps, $\mathcal{F}$: Number of Fine-tuning steps, $PCR(x, y)$: Reward Model (Phoneme Count Ratio), $\delta$ : list of $\mathcal{F}$ threshold values, $ST-Flag$ \\

    Train Model $\mathcal{M}$ on Dataset $\mathcal{D} = \{(x^{i}, y^{i})|_{i=1}^{N}\}$ using Loss $\mathcal{L}_{CE}$. \\

    \For{$g$ = $1$ to $\mathcal{G}$}{
        Generate Dataset $\mathcal{D}_{G}$ using Model $\mathcal{M}$, $\mathcal{D}_G = \{ (x^{i}, \hat{y}^{i})|_{i=1}^{N} \mid x^{i} \in \mathcal{D}, \hat{y}^{i} = \mathcal{M}(x^{i};\theta) \}$ \\
        Annotate Dataset $\mathcal{D}_{G}$ using the Reward Model $PCR(x, \hat{y})$ \\
        
        \For{$f$ = $1$ to $\mathcal{F}$}{
            Create Filtered Dataset $\mathcal{D}_F$, $\mathcal{D}_F = \{(x^i, \hat{y}^{i})|_{i=1}^{N'} \mid (x^{i}, \hat{y}^{i}) \in \mathcal{D}_G, PCR(x^i, \hat{y}^i) \in [1-\delta_f, 1+\delta_f]\}$ \\
            Train Model $\mathcal{M}$ on the Filtered Dataset $\mathcal{D}_f$ using Loss $\mathcal{L}_{CE}$
        }
    }
    \If{ST-Flag \textbf{is true}}{
        Train Model $\mathcal{M}$ on Filtered Dataset $\mathcal{D}_f$ using Loss $\mathcal{L} = \mathcal{L}_{CE} + \alpha \mathcal{L}_{KL}$
    }
    \textbf{Output: }Model $\mathcal{M}$
\end{algorithm*}
For the task of Isometric NMT, we require that the number of phonemes in the output translation of the model be as close as possible to the number of phonemes in the source sentence. In RL, the model observes the environment and takes some action. Based on this action the reward function gives some reward to the model. Then the model is trained to optimize this reward. In our approach, we use a function of the ratio between the phoneme counts in the source and target sentences as the reward equivalent.
The algorithm of our work is depicted in Alg.\ref{alg:RLNMT}. 

We first train an existing (pretrained) NMT model on a bilingual corpus to obtain $\cM$. Given a source language sentence, $\mathbf{x} = (x_1, x_2, ..., x_n)$ and the target language sentence $\mathbf{y} = (y_1, y_2, ..., y_m)$, the NMT model minimizes the Cross-Entropy Loss which is shown in Eq. \ref{eq:ce}
\begin{multline}
    \label{eq:ce}
    \small
    \mathcal{L}_{CE} = - \frac{1}{N} \sum_{i=1}^{N} \sum_{k\in V} \{\ind(y_i = k) \\ \times \log p(\hat{y}_i = k|y_{<i},\mathbf{x};\theta)\}
\end{multline}
where $N$ is the number of tokens in the output sentence, $V$ is the (output) vocabulary, $y_i$ is the $i^{th}$ word in ground-truth target language sentence and $\hat{y}_i$ is the $i^{th}$ word in the predicted target language sentence. \\
Next, we translate all the source language sentences (from the entire training corpus on which $\cM$ was trained on) using $\cM$ and obtain the output translations. This forms the generation step which corresponds to the action step in RL terminology (as shown in Fig.~\ref{fig:rl}). We compute the Phoneme Count Ratio (PCR) between the number of phonemes in the source and target sentences. This PCR acts as the reward model. We, then filter out the sentence pairs whose PCR does not lie in the specified range $[1-\delta, 1+\delta]$, where $\delta$ is the threshold. We iteratively reduce the threshold and finetune the NMT model ($\cM$) on the filtered dataset.  After this, we perform the Generation step again, using the model that is produced after the final finetuning step. Iteratively finetuning the NMT model on sentences whose PCR is closer to 1, reinforces the trained model to generate sentences matching the phoneme count of the input. Hence, One RL step consists of one Generation steps followed by multiple finetuning steps. 
 
\subsection{Proposed Student Teacher NMT Architecture (ST-RL-NMT)}

\begin{figure}[!ht]
    \centering
    \includegraphics[scale=0.8]{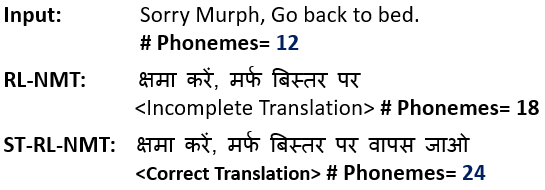}
    \vspace{-0.7cm}
    \caption{Example of quality degradation with RL-NMT and improvement achieved with ST-RL-NMT}
    \label{fig:st example}
    \vspace{-0.3cm}
\end{figure}

When we optimize the model to generate outputs where the phoneme counts in the source and target sentences are similar, we face a trade-off in the quality of the translation as shown in Fig.~\ref{fig:st example}. 
We see in the example that constraining the PCR, can sometimes lead to incomplete translations. In order to overcome this issue of quality degradation, we propose a student-teacher architecture to further finetune the trained model $\cM$, in addition to the RL approach. We use the NMT model trained on the entire parallel corpus (without the RL approach) as the \textit{teacher} model. This model produces high-quality output but the phoneme counts between the output and source sentences may not be similar. We use the RL-NMT model as the \textit{student}, which has better phoneme count compliance, but possibly, poor quality. Employing finetuning on the student model with the teacher model provides a balance between translation quality and phoneme count compliance. 
In the ST-framework, we add consistency loss term while finetuning to make the output probability distribution of the student model closer to the teacher model. We use the KL Divergence \cite{csiszár_körner_2011} between the output probability distributions of the student and teacher model as the consistency loss. This transfers the knowledge of the teacher model to the student model. We expect that, as the teacher model generates \textit{good quality} output, it will improve the \textit{quality} of the student model. Furthermore, finetuning on the phoneme count compliant parallel corpus will keep the phoneme counts of the output translations and source sentences close to each other. The KL Divergence loss term used as the consistency loss term in the training of student-teacher architecture is given in Eq. \ref{eq:kl}
\begin{equation}
    \label{eq:kl}
    \small
    \mathcal{L}_{KL} = \sum_{i=1}^{N} \mathds{KL}\left(p(*|y_{<i}, \mathbf{x}, \theta^{s})|| p(*|y_{<i}, \mathbf{x}, \theta^{t})\right)
\end{equation}
where $p(*|y_{<i}, \mathbf{x}, \theta^{s})$ represents the probability distribution of student model and $p(*|y_{<i}, \mathbf{x}, \theta^{t})$ represents the probability distribution of the teacher model.
The overall loss term used in the training of student-teacher architecture is given in Eq. \ref{eq:loss}
\begin{equation}
    \label{eq:loss}
    \mathcal{L} = \mathcal{L}_{CE} + \alpha\mathcal{L}_{KL}
\end{equation}
where $\alpha$ is a scaling factor for the KL loss ($\mathcal{L}_{KL}$).
\subsection{Proposed Phoneme Count Compliance Score} 
Previous approaches used word or character count compliance scores for evaluation, but we propose a Phoneme Count Compliance (PCC) score in this paper. The PCC score $PCC_\delta$ for a particular threshold $\delta$ denotes the percentage of sentence pairs whose phoneme count ratio (PCR) lies in the range $[1-\delta, 1+\delta]$. If $s$ denotes the phoneme count in the source sentence and $t$ denotes the phoneme count in the translated sentence then the PCR is given in Eq.~\ref{eq:pcr}, 
\begin{equation}
    \label{eq:pcr}
    PCR = s/t.
\end{equation}
If $N$ is the number of parallel sentences in the test set then the $PCC_\delta$ score is given in Eq.~\ref{eq:pcc}
\begin{multline}
    \label{eq:pcc}
    \scriptsize
    PCC_\delta = \left(\sum_{i=1}^{N} \ind[PCR(s_i, t_i) \in [1-\delta, 1+\delta]]\right) \\ \times (100 / N).    
\end{multline}
We evaluate all the models on the PCC scores for the threshold ($\delta$) values of 0.2 and 0.1. 

\indent Our primary reasons for choosing phoneme count rather than syllable count was that in Indian languages like Hindi and Marathi, there isn't a one-to-one correspondence between the letter (akshara) and syllable due to the presence of sandhi and so on. As a result, there exists multiple ways to split a word into syllables (CV, CVC, CCCV, etc.) resulting in variable syllable counts for the same sentence \cite{raj2007text,choudhury2003rule}. Hence, we believed controlling the length of the output in NMT using syllable count won't be a feasible option despite the fact that syllables have more correlation with the duration. However, we would like to mention that the PCC, although being a crude measure of length duration, is a very fast method to quickly estimate the duration of the output speech. We have taken inspirations from \cite{rasanen2021alice}, and \cite{fujita2021phoneme}. Although many other methods which explicitly estimate the time duration are available, they are computationally expensive and time-consuming \cite{wu2023videodubber}. While our approach gives reasonable results although being less \textit{nuanced}.
\section{Experiments and Results}
\subsection{Dataset}
\textbf{Training Data} \quad We use the English-Hindi parallel corpus from the Bharat Parallel Corpus Collection (BPCC) \cite{gala2023indictrans2} for training the NMT model. BPCC is a combination of various publicly available parallel corpora for 22 Indic languages, which contains human-annotated as well as automatically mined data. It contains around 39 million parallel sentences for the English-Hindi language pair, which we used for training the NMT models. We preprocess the data using the Indic NLP Library \cite{kunchukuttan2020indicnlp}

\textbf{Evaluation Data} \quad We evaluate the model on various standard test sets such as Facebook Low Resource (Flores) \cite{nllbteam2022language}, Movie subtitles, and BPCC test set. The Flores test set contains 1012 parallel sentences each for more than 200 languages in multiple domains. We focused on the English-Hindi language pair. The movie subtitles test set contains the subtitles of a Hollywood movie in English and Hindi. The BPCC test contains two parts, the general domain test set and the conversational domain test set. Both the general and conversational domain test sets contain 1023 parallel sentences. 
\subsection{Evaluation Metrics}
\textbf{BLEU} \quad BLEU \cite{papineni2002bleu} score is an automatic evaluation metric that scores the translated sentences with respect to the gold translation based on n-gram matchings. We use the Sacrebleu\footnote{\label{sacrebleu}\url{https://github.com/mjpost/sacrebleu}} implementation for generating the BLEU scores for evaluating all the models. \\
\textbf{chrF} \quad chrF \cite{popovic2015chrf} is a evaluation metric for machine translation based on character n-gram F1 scores. We use the Sacrebleu\footref{sacrebleu} implementation for generating the chrF scores for evaluating all the models. \\
\textbf{BLEURT} \quad BLEURT \cite{sellam2020bleurt} is a metric that uses a trained BERT model to evaluate the quality of the machine translation output. The model takes the reference and candidate sentence as input and outputs a score ranging from 0 to 1 based on the translation quality. \\
\textbf{COMET}\quad We also compute the COMET scores \cite{rei-etal-2020-comet} for the various models since COMET is known to correlate highly with human judgements \cite{sai-b-etal-2023-indicmt}. We use the default model, i.e., wmt22-comet-da for our experiments. This model employs a reference-based regression approach and is built upon the XLM-R architecture. It has been trained on direct assessments from WMT17 to WMT20 and provides scores ranging from 0 to 100\%, where 100\% signifies a perfect translation \cite{rei-etal-2020-comet}.
\subsection{Model Architecture}
We use the model architecture of the publicly available IndicTrans2 \cite{gala2023indictrans2} model in all our experiments. The IndicTrans2 model is based on the Transformer architecture and supports 22 Indic languages. We note that we use the default hyperparameters of the IndicTrans2 model in all our experiments. Both student and teacher network has exactly same architecture. We took one as the $\alpha$ in the distillation step in order to give equal weightage to both the student and teacher networks. We train all the models using the Nvidia A100 40GB GPU and training one model takes 30 hours on average. The detailed model parameters are shown in Table \ref{tab:model-arch}. 
\begin{table}[H]
    \centering
    \begin{tabular}{|p{11em}|c|}
        \hline
        \textbf{Parameter} & \textbf{Value} \\
        \hline
        \# encoder layers & 18 \\
        \# decoder layers & 18 \\
        \# encoder attention heads & 16 \\
        \# decoder attention heads & 16 \\
        Encoder embedding \hspace{1cm} dimensions & 1024 \\
        Decoder embedding dimensions & 1024 \\
        Encoder feedforward layer dimensions & 8192 \\
        Decoder feedforward layer dimensions & 8192 \\
        Total number of parameters & 1.1 Billion \\
        \hline
    \end{tabular}
    \caption{Details of the model architecture.}
    \label{tab:model-arch}
\end{table}
\begin{table*}[!ht]
    \centering
    \setlength{\tabcolsep}{4pt}
    \resizebox{\textwidth}{!}{%
    \begin{tabular}{|c|cccccc|cccccc|}
\hline
Model                                                           & \multicolumn{6}{c|}{Movie Test Set}                                                                                                                                                                                       & \multicolumn{6}{c|}{FLoRes}                                                                                                                                                                                               \\ \hline
\multirow{2}{*}{}                                               & \multicolumn{1}{c|}{\multirow{2}{*}{BLEU}} & \multicolumn{1}{c|}{\multirow{2}{*}{chrF}} & \multicolumn{1}{c|}{\multirow{2}{*}{BLEURT}} & \multicolumn{1}{c|}{\multirow{2}{*}{COMET}} & \multicolumn{2}{c|}{PCC}           & \multicolumn{1}{c|}{\multirow{2}{*}{BLEU}} & \multicolumn{1}{c|}{\multirow{2}{*}{chrF}} & \multicolumn{1}{c|}{\multirow{2}{*}{BLEURT}} & \multicolumn{1}{c|}{\multirow{2}{*}{COMET}} & \multicolumn{2}{c|}{PCC}           \\ \cline{6-7} \cline{12-13} & \multicolumn{1}{c|}{}                      & \multicolumn{1}{c|}{}                      & \multicolumn{1}{c|}{}                        & \multicolumn{1}{c|}{}                       & \multicolumn{1}{c|}{0.2}   & 0.1   & \multicolumn{1}{c|}{}                      & \multicolumn{1}{c|}{}                      & \multicolumn{1}{c|}{}                        & \multicolumn{1}{c|}{}                       & \multicolumn{1}{c|}{0.2}   & 0.1   \\ \hline
IndicTrans2                                                     & \multicolumn{1}{c|}{38.41}                 & \multicolumn{1}{c|}{62.02}                 & \multicolumn{1}{c|}{0.76}                    & \multicolumn{1}{c|}{84.13}                  & \multicolumn{1}{c|}{36.77} & 16.99 & \multicolumn{1}{c|}{36.35}                 & \multicolumn{1}{c|}{60.71}                 & \multicolumn{1}{c|}{0.73}                    & \multicolumn{1}{c|}{81.53}                  & \multicolumn{1}{c|}{72.72} & 36.56 \\ \hline
IndicTrans2 FT                                                  & \multicolumn{1}{c|}{42.25}                 & \multicolumn{1}{c|}{64.28}                 & \multicolumn{1}{c|}{0.76}                    & \multicolumn{1}{c|}{83.93}                  & \multicolumn{1}{c|}{38.85} & 18.46 & \multicolumn{1}{c|}{36.59}                 & \multicolumn{1}{c|}{60.72}                 & \multicolumn{1}{c|}{0.72}                    & \multicolumn{1}{c|}{80.66}                  & \multicolumn{1}{c|}{73.81} & 37.15 \\ \hline
Isometric MT                                                    & \multicolumn{1}{c|}{31.67}                 & \multicolumn{1}{c|}{57.89}                 & \multicolumn{1}{c|}{0.59}                    & \multicolumn{1}{c|}{72.52}                  & \multicolumn{1}{c|}{12.84} & 4.67  & \multicolumn{1}{c|}{31.34}                 & \multicolumn{1}{c|}{56.79}                 & \multicolumn{1}{c|}{0.67}                    & \multicolumn{1}{c|}{75.81}                  & \multicolumn{1}{c|}{58.2}  & 29.15 \\ \hline
NLLB (1.3B)                                                     & \multicolumn{1}{c|}{36.87}                 & \multicolumn{1}{c|}{59.53}                 & \multicolumn{1}{c|}{0.72}                    & \multicolumn{1}{c|}{81.91}                  & \multicolumn{1}{c|}{37.48} & 17.66 & \multicolumn{1}{c|}{30.40}                 & \multicolumn{1}{c|}{56.00}                 & \multicolumn{1}{c|}{0.71}                    & \multicolumn{1}{c|}{80.32}                  & \multicolumn{1}{c|}{73.22} & 38.93 \\ \hline
M2M-100                                                         & \multicolumn{1}{c|}{30.54}                 & \multicolumn{1}{c|}{54.81}                 & \multicolumn{1}{c|}{0.68}                    & \multicolumn{1}{c|}{76.15}                  & \multicolumn{1}{c|}{28.04} & 13.5  & \multicolumn{1}{c|}{26.18}                 & \multicolumn{1}{c|}{50.82}                 & \multicolumn{1}{c|}{0.65}                    & \multicolumn{1}{c|}{73.88}                  & \multicolumn{1}{c|}{68.67} & 34.18 \\ \hline
LLaMA2-7B                                                       & \multicolumn{1}{c|}{29.27}                 & \multicolumn{1}{c|}{50.66}                 & \multicolumn{1}{c|}{0.663}                        & \multicolumn{1}{c|}{77.73}                  & \multicolumn{1}{c|}{0.45}  & 0.22  & \multicolumn{1}{c|}{17.48}                 & \multicolumn{1}{c|}{39.55}                 & \multicolumn{1}{c|}{0.512}                        & \multicolumn{1}{c|}{62.49}                  & \multicolumn{1}{c|}{0.39}  & 0.24  \\ \hline
\begin{tabular}[c]{@{}c@{}}RL-NMT \\  (Proposed)\end{tabular}   & \multicolumn{1}{c|}{32.73}                 & \multicolumn{1}{c|}{54.19}                 & \multicolumn{1}{c|}{0.71}                    & \multicolumn{1}{c|}{80.05}                  & \multicolumn{1}{c|}{72.14} & 39.32 & \multicolumn{1}{c|}{34.31}                 & \multicolumn{1}{c|}{58.75}                 & \multicolumn{1}{c|}{0.71}                    & \multicolumn{1}{c|}{79.70}                  & \multicolumn{1}{c|}{91.3}  & 50.39 \\ \hline
\begin{tabular}[c]{@{}c@{}}ST-RL-NMT \\ (Proposed)\end{tabular} & \multicolumn{1}{c|}{37.70}                 & \multicolumn{1}{c|}{59.03}                 & \multicolumn{1}{c|}{0.73}                    & \multicolumn{1}{c|}{81.85}                  & \multicolumn{1}{c|}{58.92} & 30.03 & \multicolumn{1}{c|}{35.67}                 & \multicolumn{1}{c|}{60.07}                 & \multicolumn{1}{c|}{0.72}                    & \multicolumn{1}{c|}{80.34}                  & \multicolumn{1}{c|}{81.52} & 42.58 \\ \hline
\end{tabular}
}
    \caption{Results of evaluation of different models on BLEU, chrF, BLEURT, COMET and PCC scores on the Movie and FLoRes test set. The scores reported are the average values obtained.}
    \label{tab:results-1}
\end{table*}
\begin{table*}[!ht]
    \centering
    \setlength{\tabcolsep}{4pt}
    \resizebox{\textwidth}{!}{%
    \begin{tabular}{|c|cccccc|cccccc|}
\hline
Model                                                           & \multicolumn{6}{c|}{BPCC General}                                                                                                                                                                                         & \multicolumn{6}{c|}{BPCC Conversational}                                                                                                                                                                                  \\ \hline
\multirow{2}{*}{}                                               & \multicolumn{1}{c|}{\multirow{2}{*}{BLEU}} & \multicolumn{1}{c|}{\multirow{2}{*}{chrF}} & \multicolumn{1}{c|}{\multirow{2}{*}{BLEURT}} & \multicolumn{1}{c|}{\multirow{2}{*}{COMET}} & \multicolumn{2}{c|}{PCC}           & \multicolumn{1}{c|}{\multirow{2}{*}{BLEU}} & \multicolumn{1}{c|}{\multirow{2}{*}{chrF}} & \multicolumn{1}{c|}{\multirow{2}{*}{BLEURT}} & \multicolumn{1}{c|}{\multirow{2}{*}{COMET}} & \multicolumn{2}{c|}{PCC}           \\ \cline{6-7} \cline{12-13} 
& \multicolumn{1}{c|}{}                      & \multicolumn{1}{c|}{}                      & \multicolumn{1}{c|}{}                        & \multicolumn{1}{c|}{}                       & \multicolumn{1}{c|}{0.2}   & 0.1   & \multicolumn{1}{c|}{}                      & \multicolumn{1}{c|}{}                      & \multicolumn{1}{c|}{}                        & \multicolumn{1}{c|}{}                       & \multicolumn{1}{c|}{0.2}   & 0.1   \\ \hline
IndicTrans2                                                     & \multicolumn{1}{c|}{32.57}                 & \multicolumn{1}{c|}{57.87}                 & \multicolumn{1}{c|}{0.72}                    & \multicolumn{1}{c|}{80.45}                  & \multicolumn{1}{c|}{81.25} & 45.31 & \multicolumn{1}{c|}{28.55}                 & \multicolumn{1}{c|}{49.85}                 & \multicolumn{1}{c|}{0.76}                    & \multicolumn{1}{c|}{80.45}                  & \multicolumn{1}{c|}{46.5}  & 23.55 \\ \hline
IndicTrans2 FT                                                  & \multicolumn{1}{c|}{27.94}                 & \multicolumn{1}{c|}{55.29}                 & \multicolumn{1}{c|}{0.70}                    & \multicolumn{1}{c|}{79.01}                  & \multicolumn{1}{c|}{81.93} & 46.97 & \multicolumn{1}{c|}{26.83}                 & \multicolumn{1}{c|}{48.84}                 & \multicolumn{1}{c|}{0.75}                    & \multicolumn{1}{c|}{79.01}                  & \multicolumn{1}{c|}{49.76} & 24.68 \\ \hline
Isometric MT                                                    & \multicolumn{1}{c|}{23.77}                 & \multicolumn{1}{c|}{50.79}                 & \multicolumn{1}{c|}{0.65}                    & \multicolumn{1}{c|}{73.31}                  & \multicolumn{1}{c|}{70.31} & 39.06 & \multicolumn{1}{c|}{20.97}                 & \multicolumn{1}{c|}{45.51}                 & \multicolumn{1}{c|}{0.70}                    & \multicolumn{1}{c|}{78.72}                  & \multicolumn{1}{c|}{31.13} & 13.63 \\ \hline
NLLB (1.3B)                                                     & \multicolumn{1}{c|}{25.41}                 & \multicolumn{1}{c|}{53.01}                 & \multicolumn{1}{c|}{0.706}                   & \multicolumn{1}{c|}{78.93}                  & \multicolumn{1}{c|}{80.6}  & 46.5  & \multicolumn{1}{c|}{25.89}                 & \multicolumn{1}{c|}{47.57}                 & \multicolumn{1}{c|}{0.75}                    & \multicolumn{1}{c|}{83.75}                  & \multicolumn{1}{c|}{47.77} & 24.22 \\ \hline
LLaMA2-7B                                                       & \multicolumn{1}{c|}{10.35}                 & \multicolumn{1}{c|}{31.98}                 & \multicolumn{1}{c|}{0.532}                        & \multicolumn{1}{c|}{61.25}                  & \multicolumn{1}{c|}{0.34}  & 0.2   & \multicolumn{1}{c|}{17.23}                 & \multicolumn{1}{c|}{43.02}                 & \multicolumn{1}{c|}{0.701}                        & \multicolumn{1}{c|}{80.5}                   & \multicolumn{1}{c|}{0.49}  & 0.23  \\ \hline
M2M-100                                                         & \multicolumn{1}{c|}{18.15}                 & \multicolumn{1}{c|}{44.62}                 & \multicolumn{1}{c|}{0.63}                    & \multicolumn{1}{c|}{71.79}                  & \multicolumn{1}{c|}{76.95} & 43.75 & \multicolumn{1}{c|}{17.17}                 & \multicolumn{1}{c|}{40.01}                 & \multicolumn{1}{c|}{0.69}                    & \multicolumn{1}{c|}{78.10}                  & \multicolumn{1}{c|}{43.24} & 20.69 \\ \hline
\begin{tabular}[c]{@{}c@{}}RL-NMT \\ (Proposed)\end{tabular}    & \multicolumn{1}{c|}{27.26}                 & \multicolumn{1}{c|}{54.67}                 & \multicolumn{1}{c|}{0.7045}                  & \multicolumn{1}{c|}{78.33}                  & \multicolumn{1}{c|}{92.38} & 58.10 & \multicolumn{1}{c|}{24.55}                 & \multicolumn{1}{c|}{46.45}                 & \multicolumn{1}{c|}{0.73}                    & \multicolumn{1}{c|}{78.33}                  & \multicolumn{1}{c|}{83.23} & 48.10 \\ \hline
\begin{tabular}[c]{@{}c@{}}ST-RL-NMT \\ (Proposed)\end{tabular} & \multicolumn{1}{c|}{27.65}                 & \multicolumn{1}{c|}{55.12}                 & \multicolumn{1}{c|}{0.7091}                  & \multicolumn{1}{c|}{78.80}                  & \multicolumn{1}{c|}{87.20} & 53.71 & \multicolumn{1}{c|}{25.98}                 & \multicolumn{1}{c|}{47.88}                 & \multicolumn{1}{c|}{0.75}                    & \multicolumn{1}{c|}{78.80}                  & \multicolumn{1}{c|}{67.07} & 34.39 \\ \hline
\end{tabular}}
    \caption{Results of evaluation of different models on BLEU, chrF, BLEURT, COMET and PCC scores on the BPCC test set. The scores reported are the average values obtained.}
    \label{tab:results-2}
\end{table*}
\begin{figure*}[!ht]
    \centering
    \includegraphics[width=\textwidth]{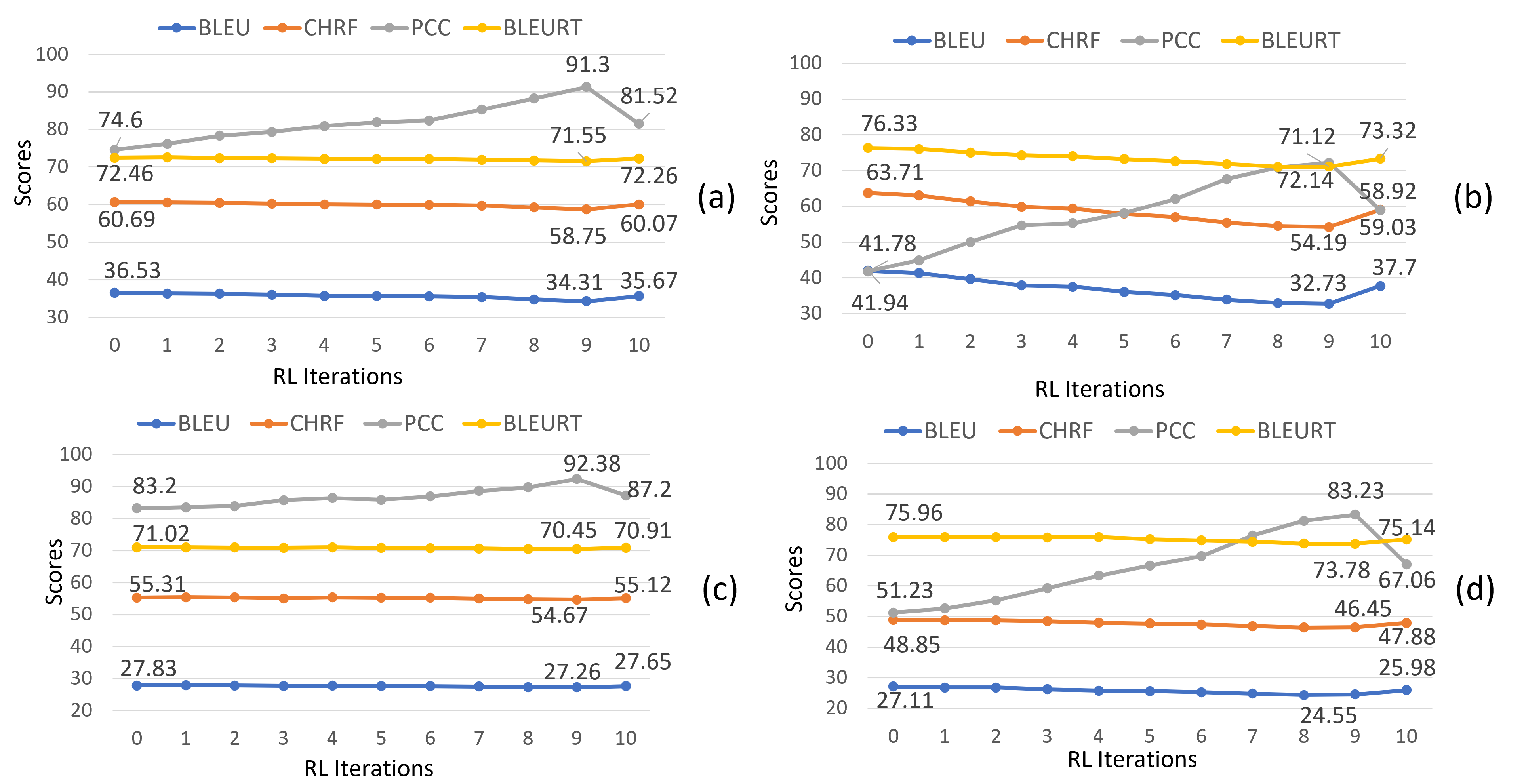}
    \caption{Plot showing the different evaluation metrics at each RL-Step for (a) FLoRes, (b) Movie, (C) BPCC General and (d) BPCC Conversational Tests. Here, last step is with the student-teacher objective.}
    \label{fig:scores}
\end{figure*}
\subsection{Baselines}
We compare our approach with various state-of-the-art models given as follows. \\
\textbf{IndicTrans2} \quad We compare our approach with the SOTA IndicTrans2 \cite{gala2023indictrans2} model without applying any phoneme count control measures.\\
\textbf{IndicTrans2-FT} \quad We fine-tune (FT) IndicTrans2 model using only English-Hindi data in order to improve the performance of SOTA model for the selected language pair. This model achieves highest BLEU score, i.e., it performs best w.r.t. quality of translation. Hence, the same model is selected for Teacher in our proposed architecture.\\
\textbf{Isometric MT} \quad The Isometric MT \cite{lakew2022isometric} approach controls the number of words generated in the translation. The source language sentences are tagged with '<short>', '<normal>' or '<long>' tag depending on the ratio between the word counts in the source and target language sentences. During inference, the input sentences are tagged with the '<normal>' tag to generate length compliant sentences. \\
\textbf{No Language Left Behind (NLLB)} \quad NLLB \cite{nllbteam2022language} is a state-of-the-art multilingual NMT model that can translate among 200 languages. The NLLB model makes use of a sparse mixture of expert models with shared and specialized capacity to improve the performance of low-resource languages. The NLLB model also makes use of large-scale data augmentation with back-translation. The distilled NLLB model has 1.3 billion parameters in total. \\
\textbf{M2M-100} \quad M2M-100 \cite{fan2021beyond} is a multilingual NMT model that can translate among 100 languages. The M2M-100 model is trained on a parallel corpus of 2,200 language directions without relying on English-centric datasets. The M2M-100 model gives good performance improvements over bilingual NMT models. The M2M-100 model has 418 million parameters in total. \\
\textbf{LLaMA2 \& LLaMA2-FT} \quad LLaMA-2 \cite{touvron2023LLaMA} is a large language model (LLM) with 7 billion parameters and is trained on 2 trillion tokens. The baseline LLaMA-2 model did not give good performance for the English-Hindi translation task, so we finetuned (FT) the LLaMA-2 model on 1 million randomly sampled sentence pairs from the training set of BPCC corpus. 

\subsection{Results}\label{sec:results}
\begin{figure}[H]
  \centering
\includegraphics[width=0.5\textwidth]{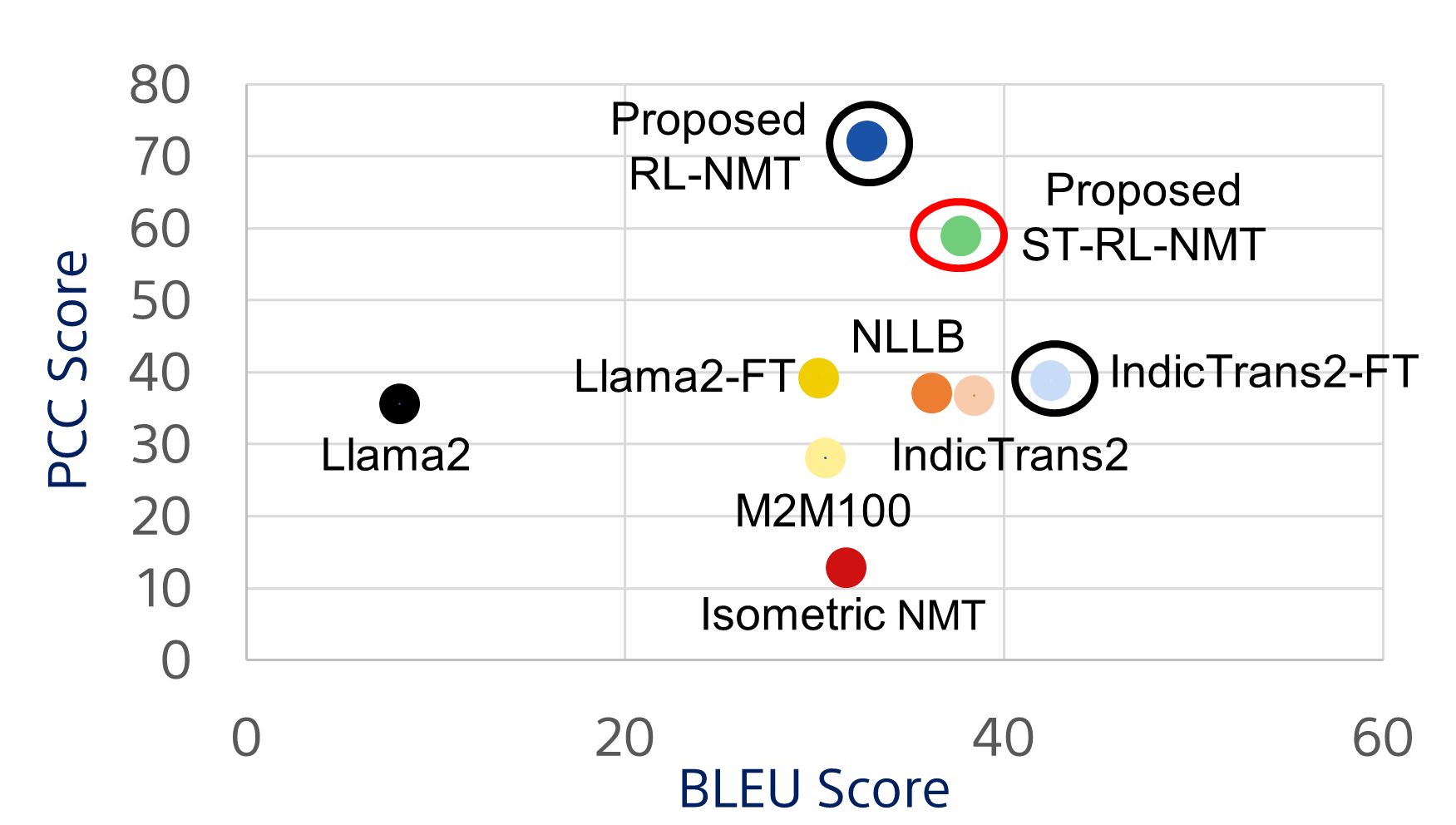}
  \vspace{-0.5cm}
  \caption{Trade-off between BLEU score \textit{vs.} PCC score}
  \label{Fig:Tradeoff}
  \vspace{-0.5cm}
\end{figure}

\begin{figure}[H]
    \centering
  \includegraphics[width=0.5\textwidth]{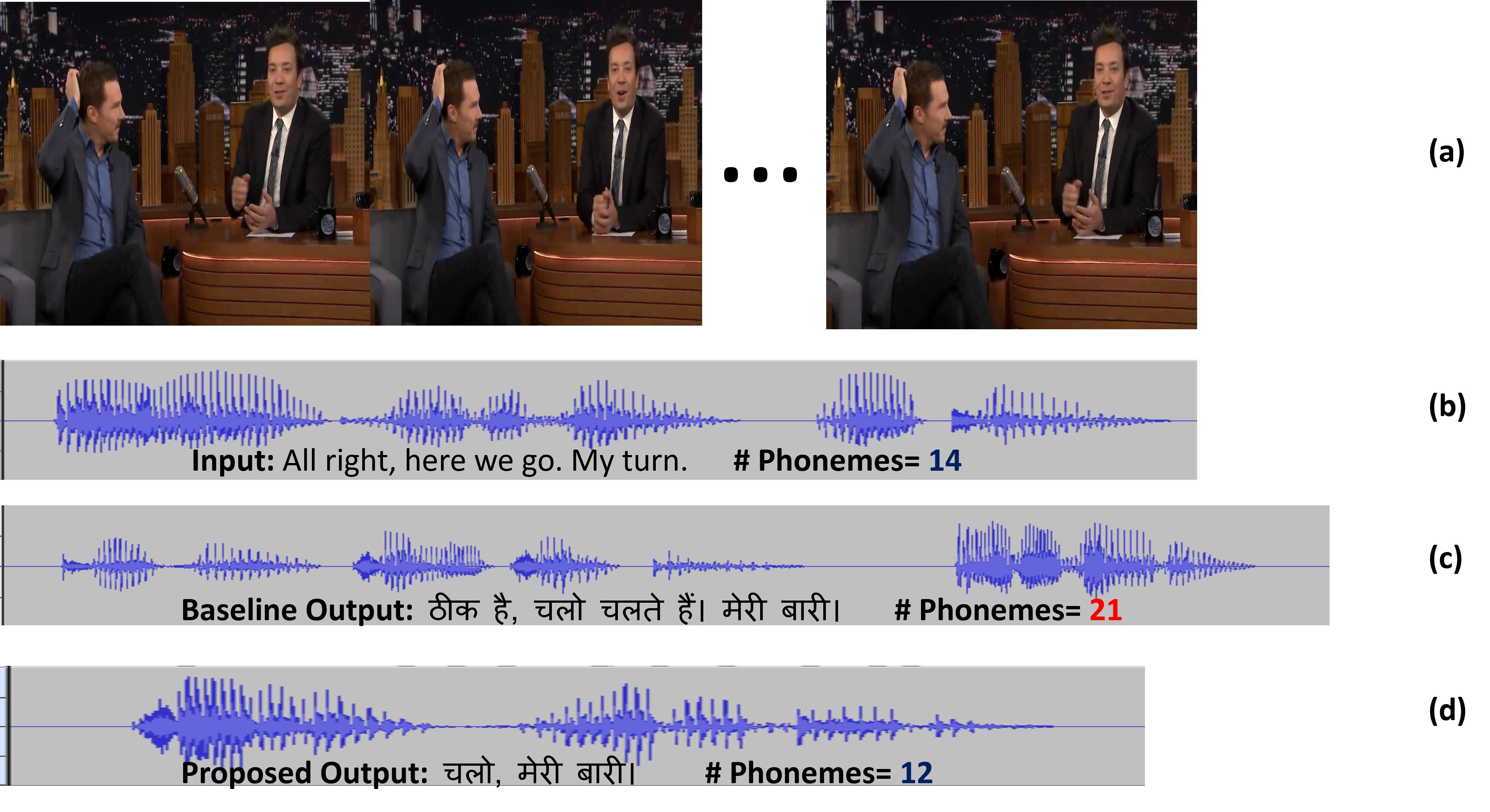}
  \vspace{-0.7cm}
\caption{A qualitative example of AVD using baseline and proposed approach.}
\vspace{-0.3cm}
\label{fig:eg1}
\end{figure}

Table \ref{tab:results-1} and Table \ref{tab:results-2} present results obtained using various SOTA as well as proposed approaches on four test sets, namely, Movie, FLoRes, BPCC General and BPCC Conversational corpora. We see significant improvements in PCC values for both the p=0.2 and p=0.1 cases. Specifically, the proposed RL-NMT technique has attained absolute improvements ranging from 10\% to 33\% in PCC values across the various evaluation test sets. However, on the contrary, there has been an observed absolute decrease of 2\% to 10\% in BLEU scores, chrF scores, COMET scores and BLEURT scores, which primarily indicate the quality of translation. Furthermore, based on Table \ref{tab:results-1} and Table \ref{tab:results-2}, it can be discerned that the proposed ST-RL-NMT framework is instrumental in mitigating the degradation occurring on the translation front. In particular, the proposed ST-RL-NMT framework has successfully reduced the absolute degradation in quality-related metrics from 0.5\% to 5\% compared to the previous range of 2\% to 10\% with the RL-NMT approach. This trade-off between the BLEU score and Phoneme Count Compliance Score (PCC) is visually represented in Fig. \ref{Fig:Tradeoff}. The results are presented across different baselines including LLMs (Llama2 and Llama2-FT model). It can be clearly seen that the IndicTrans2-FT model achieves the highest BLEU score. Hence, we select IndicTrans2-FT as the teacher in our proposed ST-RL-NMT approach. While the RL-NMT approach attains the best PCC scores, it does come at the expense of a decline in performance on the BLEU score side. On the other hand, the ST-RL-NMT framework is able to simultaneously achieve better trade-offs for PCC and BLEU score compared to other SOTA algorithms.\\
\indent Fig. \ref{fig:scores} presents the detailed analysis of results at each RL step during the training for all four evaluation sets. We can see that with each RL step, the PCC score is increasing significantly. On the other hand, BLEU score, BLEURT score, COMET score and chrF values are decreasing. Nevertheless, at iteration 10, where the Student Teacher framework was introduced, noticeable improvements in the BLEU, BLEURT, COMET and chrF scores can be observed. 
There can be many ways to choose which model to push for the ST post-processing. To ensure a fair comparison, we implemented the ST post-processing at the tenth iteration of the RL algorithm. However, we can plot the max-normalized BLEU scores and PCC scores and select the point where the two plots either intersect or have minimum distance. Subsequently, this model can be considered as the student and we believe that it will have a balanced compromise between the quality and length compliance. 
Fig. \ref{fig:eg1} illustrates the qualitative output generated by the baseline IndicTrans2 model and proposed ST-RL-NMT model. It is evident that the original English sentence contains 10 phonemes. Conversely, the baseline model has produced a correct translation with 18 phonemes, as it did not take into account any length-based constraints. In contrast, the proposed ST-RL-NMT model produces the correct output that adheres to the desired length, containing 10 phonemes. This results in (near-) equal duration when synthesized using the target language TTS. Therefore, minimal post-processing is required to adjust the duration in the final dubbed output. 
\section{Summary and Conclusion}
In this paper, we proposed a reinforcement learning-based training strategy for Isometric NMT. We proposed to match the count of phonemes for this task, as phonemes have a strong correlation with speech duration. Further, we enhanced our agent in the RL-based training strategy with a student-teacher architecture to circumvent the problem of quality degradation that arises from optimizing the model for generating phoneme count compliant sentences. We also proposed the Phoneme Count Compliance score to evaluate the performance of Isometric NMT models. Experimental results showed that our approach gives significant performance improvements in terms of Phoneme Compliance Scores over various state-of-the-art NMT models including LLMs.
\section*{Limitations}
In the future, on the technical front, we will investigate a \textit{soft-}threshold approach for filtering the data based on PCC and the BLEU score. On the computational front, we note that the generation step in our approach is expensive as we need to translate the entire source side of the parallel corpus. Also, we plan to perform experiments with various language pairs from different language families.
\section*{Ethics Statement}
The aim of our work is to improve the performance of NMT models for the Isometric NMT task. The datasets that we used in this work are publicly available. Publicly available datasets can contain biased sentences. We train the NMT models on the available parallel corpus, evaluate the models, and have cited the appropriate sources.

\bibliography{custom}

\begin{thebibliography}{27}
\expandafter\ifx\csname natexlab\endcsname\relax\def\natexlab#1{#1}\fi

\bibitem[{Bahdanau et~al.(2014)Bahdanau, Cho, and Bengio}]{bahdanau2014neural}
Dzmitry Bahdanau, Kyunghyun Cho, and Yoshua Bengio. 2014.
\newblock Neural machine translation by jointly learning to align and translate.
\newblock \emph{arXiv preprint arXiv:1409.0473}.

\bibitem[{Cho et~al.(2014)Cho, van Merri{\"e}nboer, Bahdanau, and Bengio}]{cho-etal-2014-properties}
Kyunghyun Cho, Bart van Merri{\"e}nboer, Dzmitry Bahdanau, and Yoshua Bengio. 2014.
\newblock \href {https://doi.org/10.3115/v1/W14-4012} {On the properties of neural machine translation: Encoder{--}decoder approaches}.
\newblock In \emph{Proceedings of {SSST}-8, Eighth Workshop on Syntax, Semantics and Structure in Statistical Translation}, pages 103--111, Doha, Qatar. Association for Computational Linguistics.

\bibitem[{Choudhury(2003)}]{choudhury2003rule}
Monojit Choudhury. 2003.
\newblock Rule-based grapheme to phoneme mapping for hindi speech synthesis.
\newblock In \emph{90th Indian Science Congress of the International Speech Communication Association (ISCA), Bangalore, India}. Citeseer.

\bibitem[{Csiszár and Körner(2011)}]{csiszár_körner_2011}
Imre Csiszár and János Körner. 2011.
\newblock \href {https://doi.org/10.1017/CBO9780511921889} {\emph{Information Theory: Coding Theorems for Discrete Memoryless Systems}}, 2 edition.
\newblock Cambridge University Press.

\bibitem[{Fan et~al.(2021)Fan, Bhosale, Schwenk, Ma, El-Kishky, Goyal, Baines, Celebi, Wenzek, Chaudhary et~al.}]{fan2021beyond}
Angela Fan, Shruti Bhosale, Holger Schwenk, Zhiyi Ma, Ahmed El-Kishky, Siddharth Goyal, Mandeep Baines, Onur Celebi, Guillaume Wenzek, Vishrav Chaudhary, et~al. 2021.
\newblock Beyond english-centric multilingual machine translation.
\newblock \emph{Journal of Machine Learning Research}, 22(107):1--48.

\bibitem[{Fujita et~al.(2021)Fujita, Ando, and Ijima}]{fujita2021phoneme}
Kenichi Fujita, Atsushi Ando, and Yusuke Ijima. 2021.
\newblock Phoneme duration modeling using speech rhythm-based speaker embeddings for multi-speaker speech synthesis.
\newblock In \emph{Interspeech}, pages 3141--3145.

\bibitem[{Gala et~al.(2023)Gala, Chitale, AK, Doddapaneni, , Kumar, Nawale, Sujatha, Puduppully, Raghavan, Kumar, Khapra, Dabre, and Kunchukuttan}]{gala2023indictrans2}
Jay Gala, Pranjal~A. Chitale, Raghavan AK, Varun Gumma~Sumanth Doddapaneni, , Aswanth Kumar, Janki Nawale, Anupama Sujatha, Ratish Puduppully, Vivek Raghavan, Pratyush Kumar, Mitesh~M. Khapra, Raj Dabre, and Anoop Kunchukuttan. 2023.
\newblock \href {https://openreview.net/forum?id=vfT4YuzAYA} {Indictrans2: Towards high-quality and accessible machine translation models for all 22 scheduled indian languages}.
\newblock \emph{Transactions on Machine Learning Research}.

\bibitem[{Gulcehre et~al.(2023)Gulcehre, Paine, Srinivasan, Konyushkova, Weerts, Sharma, Siddhant, Ahern, Wang, Gu, Macherey, Doucet, Firat, and de~Freitas}]{ReSTDeepMind}
Caglar Gulcehre, Tom~Le Paine, Srivatsan Srinivasan, Ksenia Konyushkova, Lotte Weerts, Abhishek Sharma, Aditya Siddhant, Alex Ahern, Miaosen Wang, Chenjie Gu, Wolfgang Macherey, Arnaud Doucet, Orhan Firat, and Nando de~Freitas. 2023.
\newblock \href {http://arxiv.org/abs/2308.08998} {Reinforced self-training (rest) for language modeling}.

\bibitem[{Hinton et~al.(2015)Hinton, Vinyals, and Dean}]{hinton2015distilling}
Geoffrey Hinton, Oriol Vinyals, and Jeff Dean. 2015.
\newblock Distilling the knowledge in a neural network.
\newblock \emph{arXiv preprint arXiv:1503.02531}.

\bibitem[{Kunchukuttan(2020)}]{kunchukuttan2020indicnlp}
Anoop Kunchukuttan. 2020.
\newblock {The IndicNLP Library}.
\newblock \url{https://github.com/anoopkunchukuttan/indic_nlp_library/blob/master/docs/indicnlp.pdf}.

\bibitem[{Lakew et~al.(2021)Lakew, Federico, Wang, Hoang, Virkar, Barra-Chicote, and Enyedi}]{lakew21verbosity}
Surafel~M. Lakew, Marcello Federico, Yue Wang, Cuong Hoang, Yogesh Virkar, Roberto Barra-Chicote, and Robert Enyedi. 2021.
\newblock \href {https://doi.org/10.1109/ICASSP39728.2021.9414411} {Machine translation verbosity control for automatic dubbing}.
\newblock In \emph{ICASSP 2021 - 2021 IEEE International Conference on Acoustics, Speech and Signal Processing (ICASSP)}, pages 7538--7542.

\bibitem[{Lakew et~al.(2022)Lakew, Virkar, Mathur, and Federico}]{lakew2022isometric}
Surafel~M Lakew, Yogesh Virkar, Prashant Mathur, and Marcello Federico. 2022.
\newblock Isometric mt: Neural machine translation for automatic dubbing.
\newblock In \emph{ICASSP 2022-2022 IEEE International Conference on Acoustics, Speech and Signal Processing (ICASSP)}, pages 6242--6246. IEEE.

\bibitem[{Lakew et~al.(2019)Lakew, Di~Gangi, and Federico}]{lakew-etal-2019-controlling}
Surafel~Melaku Lakew, Mattia Di~Gangi, and Marcello Federico. 2019.
\newblock \href {https://aclanthology.org/2019.iwslt-1.31} {Controlling the output length of neural machine translation}.
\newblock In \emph{Proceedings of the 16th International Conference on Spoken Language Translation}, Hong Kong. Association for Computational Linguistics.

\bibitem[{Oppenheim et~al.(1999)Oppenheim, Schafer, and Buck}]{oppenheim99}
Alan~V. Oppenheim, Ronald~W. Schafer, and John~R. Buck. 1999.
\newblock \emph{Discrete-Time Signal Processing}, second edition.
\newblock Prentice-hall Englewood Cliffs.

\bibitem[{Papineni et~al.(2002)Papineni, Roukos, Ward, and Zhu}]{papineni2002bleu}
Kishore Papineni, Salim Roukos, Todd Ward, and Wei-Jing Zhu. 2002.
\newblock Bleu: a method for automatic evaluation of machine translation.
\newblock In \emph{Proceedings of the 40th annual meeting of the Association for Computational Linguistics}, pages 311--318.

\bibitem[{Popovi{\'c}(2015)}]{popovic2015chrf}
Maja Popovi{\'c}. 2015.
\newblock chrf: character n-gram f-score for automatic mt evaluation.
\newblock In \emph{Proceedings of the tenth workshop on statistical machine translation}, pages 392--395.

\bibitem[{Quatieri(2001)}]{quatieri}
Thomas Quatieri. 2001.
\newblock \emph{Discrete-Time Speech Signal Processing: Principles and Practice}, first edition.
\newblock Prentice Hall Press, USA.

\bibitem[{Raj et~al.(2007)Raj, Sarkar, Pammi, Yuvaraj, Bansal, Prahallad, and Black}]{raj2007text}
Anand~Arokia Raj, Tanuja Sarkar, Sathish~Chandra Pammi, Santhosh Yuvaraj, Mohit Bansal, Kishore Prahallad, and Alan~W Black. 2007.
\newblock Text processing for text-to-speech systems in indian languages.
\newblock In \emph{Ssw}, pages 188--193.

\bibitem[{R{\"a}s{\"a}nen et~al.(2021)R{\"a}s{\"a}nen, Seshadri, Lavechin, Cristia, and Casillas}]{rasanen2021alice}
Okko R{\"a}s{\"a}nen, Shreyas Seshadri, Marvin Lavechin, Alejandrina Cristia, and Marisa Casillas. 2021.
\newblock Alice: An open-source tool for automatic measurement of phoneme, syllable, and word counts from child-centered daylong recordings.
\newblock \emph{Behavior Research Methods}, 53:818--835.

\bibitem[{Rei et~al.(2020)Rei, Stewart, Farinha, and Lavie}]{rei-etal-2020-comet}
Ricardo Rei, Craig Stewart, Ana~C Farinha, and Alon Lavie. 2020.
\newblock \href {https://doi.org/10.18653/v1/2020.emnlp-main.213} {{COMET}: A neural framework for {MT} evaluation}.
\newblock In \emph{Proceedings of the 2020 Conference on Empirical Methods in Natural Language Processing (EMNLP)}, pages 2685--2702, Online. Association for Computational Linguistics.

\bibitem[{Sai~B et~al.(2023)Sai~B, Dixit, Nagarajan, Kunchukuttan, Kumar, Khapra, and Dabre}]{sai-b-etal-2023-indicmt}
Ananya Sai~B, Tanay Dixit, Vignesh Nagarajan, Anoop Kunchukuttan, Pratyush Kumar, Mitesh~M. Khapra, and Raj Dabre. 2023.
\newblock \href {https://doi.org/10.18653/v1/2023.acl-long.795} {{I}ndic{MT} eval: A dataset to meta-evaluate machine translation metrics for {I}ndian languages}.
\newblock In \emph{Proceedings of the 61st Annual Meeting of the Association for Computational Linguistics (Volume 1: Long Papers)}, pages 14210--14228, Toronto, Canada. Association for Computational Linguistics.

\bibitem[{Sellam et~al.(2020)Sellam, Das, and Parikh}]{sellam2020bleurt}
Thibault Sellam, Dipanjan Das, and Ankur Parikh. 2020.
\newblock Bleurt: Learning robust metrics for text generation.
\newblock In \emph{Proceedings of the 58th Annual Meeting of the Association for Computational Linguistics}, pages 7881--7892.

\bibitem[{Sutskever et~al.(2014)Sutskever, Vinyals, and Le}]{sutskever2014sequence}
Ilya Sutskever, Oriol Vinyals, and Quoc~V Le. 2014.
\newblock Sequence to sequence learning with neural networks.
\newblock \emph{Advances in neural information processing systems}, 27.

\bibitem[{Team et~al.(2022)Team, Costa-jussà, Cross, Çelebi, Elbayad, Heafield, Heffernan, Kalbassi, Lam, Licht, Maillard, Sun, Wang, Wenzek, Youngblood, Akula, Barrault, Gonzalez, Hansanti, Hoffman, Jarrett, Sadagopan, Rowe, Spruit, Tran, Andrews, Ayan, Bhosale, Edunov, Fan, Gao, Goswami, Guzmán, Koehn, Mourachko, Ropers, Saleem, Schwenk, and Wang}]{nllbteam2022language}
NLLB Team, Marta~R. Costa-jussà, James Cross, Onur Çelebi, Maha Elbayad, Kenneth Heafield, Kevin Heffernan, Elahe Kalbassi, Janice Lam, Daniel Licht, Jean Maillard, Anna Sun, Skyler Wang, Guillaume Wenzek, Al~Youngblood, Bapi Akula, Loic Barrault, Gabriel~Mejia Gonzalez, Prangthip Hansanti, John Hoffman, Semarley Jarrett, Kaushik~Ram Sadagopan, Dirk Rowe, Shannon Spruit, Chau Tran, Pierre Andrews, Necip~Fazil Ayan, Shruti Bhosale, Sergey Edunov, Angela Fan, Cynthia Gao, Vedanuj Goswami, Francisco Guzmán, Philipp Koehn, Alexandre Mourachko, Christophe Ropers, Safiyyah Saleem, Holger Schwenk, and Jeff Wang. 2022.
\newblock \href {http://arxiv.org/abs/2207.04672} {No language left behind: Scaling human-centered machine translation}.

\bibitem[{Touvron et~al.(2023)Touvron, Martin, Stone, Albert, Almahairi, Babaei, Bashlykov, Batra, Bhargava, Bhosale et~al.}]{touvron2023LLaMA}
Hugo Touvron, Louis Martin, Kevin Stone, Peter Albert, Amjad Almahairi, Yasmine Babaei, Nikolay Bashlykov, Soumya Batra, Prajjwal Bhargava, Shruti Bhosale, et~al. 2023.
\newblock Llama 2: Open foundation and fine-tuned chat models.
\newblock \emph{arXiv preprint arXiv:2307.09288}.

\bibitem[{Vaswani et~al.(2017)Vaswani, Shazeer, Parmar, Uszkoreit, Jones, Gomez, Kaiser, and Polosukhin}]{vaswani2017attention}
Ashish Vaswani, Noam Shazeer, Niki Parmar, Jakob Uszkoreit, Llion Jones, Aidan~N Gomez, {\L}ukasz Kaiser, and Illia Polosukhin. 2017.
\newblock Attention is all you need.
\newblock \emph{Advances in neural information processing systems}, 30.

\bibitem[{Wu et~al.(2023)Wu, Guo, Tan, Zhang, Li, Song, He, Zhao, Menezes, and Bian}]{wu2023videodubber}
Yihan Wu, Junliang Guo, Xu~Tan, Chen Zhang, Bohan Li, Ruihua Song, Lei He, Sheng Zhao, Arul Menezes, and Jiang Bian. 2023.
\newblock Videodubber: Machine translation with speech-aware length control for video dubbing.
\newblock In \emph{Proceedings of the AAAI Conference on Artificial Intelligence}, volume~37, pages 13772--13779.

\end{thebibliography}

\appendix
\section{Appendix}
\subsection{Machine Translation  as a Reinforcement Learning Problem}
\label{sec:appendix}
We cast the machine translation task as a Reinforcement Learning problem.  Let the input and output language vocabularies, after the suitable embeddings be denoted as $\cV_I$ and $\cV_O$, respectively. In this terminology any input (output) sentence of length $M$ will be from the finite Cartesian product $\cV_I^M(\cV_O^M)$. Hence any possible input sentence $x$ will be from the set $\bigcup\limits_{M\geq 1}\cV_I^{M}$. The output (generated) sentence $y$ will be similarly from the set $\bigcup\limits_{M\geq 1}\cV_O^M$. Also, let the distribution of the training inputs be denoted as $\cD$. We frame the problem as a Markov Decision Process (MDP) $\cM\left(\cS,\cA,\cP,r, \gamma \right)$. The state space $\cS$ is the set of all possible such tuple of vectors $(x,y)$. The action set $\cA\equiv\cV_O$. In this case, the transition kernel dynamics $\cP:\cS\times\cA\to\cS$ is defined in the following way. At any time $t$, we choose $\cP\left[(x, y_{1:t-1},a)| (x, y_{t-1}), a_t\right]=1$ if $a==a_t$, else it is 0. This makes the transition kernel deterministic. The discount parameter $\gamma$ is identically set to 0. The reward $r(.,.)$ is a function which takes two arguments, namely, $\hat{y}$ and $x$, where $\hat{y}$ is the translated sentence for the input sentence $x$ by the system. Then $r(\hat{y},x)$ is chosen as a function of the Phoneme Count Ratio (PCR) score. In particular, we set,
\[ r(\hat{y},x):=\ind\left\{PCR(\hat{y},x)\in [1-\delta,1+\delta]\right\}.\]
We impose an even stricter notion of reward for the experiments, in that we only allow sentence pairs $(x,\hat{y})$ which admit a positive reward, to be used in the fine-tuning step, and reject the zero-reward sentence pairs. We note here that the reward, as defined here, is only generated at the end of the translation of the full sentence. This notion of reward function indirectly enforces better quality translations as well as forces the output translations to adhere to strict length constraints which is essential for the automatic dubbing application. 

\end{document}